\documentclass[conference]{IEEEtran}
\IEEEoverridecommandlockouts

\usepackage{cite}
\usepackage{amsmath,amssymb,amsfonts}
\usepackage{algorithmic}
\usepackage{graphicx}
\usepackage{textcomp}
\usepackage{xcolor}
\usepackage{subcaption}
\usepackage{graphicx}
\usepackage{booktabs}
\usepackage{multirow}
\def\BibTeX{{\rm B\kern-.05em{\sc i\kern-.025em b}\kern-.08em
    T\kern-.1667em\lower.7ex\hbox{E}\kern-.125emX}}
\begin{document}

\title{SFADNet: Spatio-temporal Fused Graph based on Attention Decoupling Network for Traffic Prediction\\
\thanks{*corresponding author: sengdw@hdu.edu.cn}
}

\author{\IEEEauthorblockN{Mei Wu}
\IEEEauthorblockA{\textit{Hangzhou Dianzi University} \\
Hangzhou, China \\
222320007@hdu.edu.cn}
\and
\IEEEauthorblockN{Wenchao Weng}
\IEEEauthorblockA{\textit{Zhejiang University of Technology} \\
Hangzhou, China \\
 111124120010@zjut.edu.cn}
\and
\IEEEauthorblockN{Jun Li}
\IEEEauthorblockA{\textit{Hangzhou Dianzi University} \\
Hangzhou, China \\
232320038@hdu.edu.cn}
\and
\IEEEauthorblockN{Yiqian Lin}
\IEEEauthorblockA{\textit{Hangzhou Dianzi University} \\
Hangzhou, China \\
linyq@hdu.edu.cn}
\and
\IEEEauthorblockN{Jing Chen}
\IEEEauthorblockA{\textit{Hangzhou Dianzi University} \\
Hangzhou, China \\
cj@hdu.edu.cn}
\and
\IEEEauthorblockN{Dewen Seng*}
\IEEEauthorblockA{\textit{Hangzhou Dianzi University} \\
Hangzhou, China \\
sengdw@hdu.edu.cn}
}

\maketitle

\begin{abstract}
In recent years, traffic flow prediction has played
a crucial role in the management of intelligent transportation systems. However, traditional prediction methods are often limited by static spatial modeling, making it difficult to accurately capture the dynamic and complex relationships between time and space, thereby affecting prediction accuracy. This
paper proposes an innovative traffic flow prediction network, SFADNet, which categorizes traffic flow into multiple traffic patterns based on temporal and spatial feature matrices. For each pattern, we construct an independent adaptive spatio-temporal fusion graph based on a cross-attention mechanism, employing
residual graph convolution modules and time series modules
to better capture dynamic spatio-temporal relationships under different fine-grained traffic patterns. Extensive experimental results demonstrate that SFADNet outperforms current state-of the-art baselines across four large-scale datasets.
\end{abstract}


\section{Introduction}
The reliability and progress achieved in traffic flow prediction are propelling advancements in traffic planning, intelligent transportation systems, and urban sustainability research\cite{trafficIncidentDetection,CABIN,DeepTRANS}. While numerous conventional methods, particularly those dealing with spatial-temporal relationships, often focus solely on spatial proximity, they frequently overlook the potential temporal similarities between nodes\cite{jointPredictionModel, smoothingModel, deepLearningExperiment}.

\begin{figure}
	\centering
	\begin{subfigure}{0.24\textwidth}
		\centering
		\includegraphics[width=\linewidth]{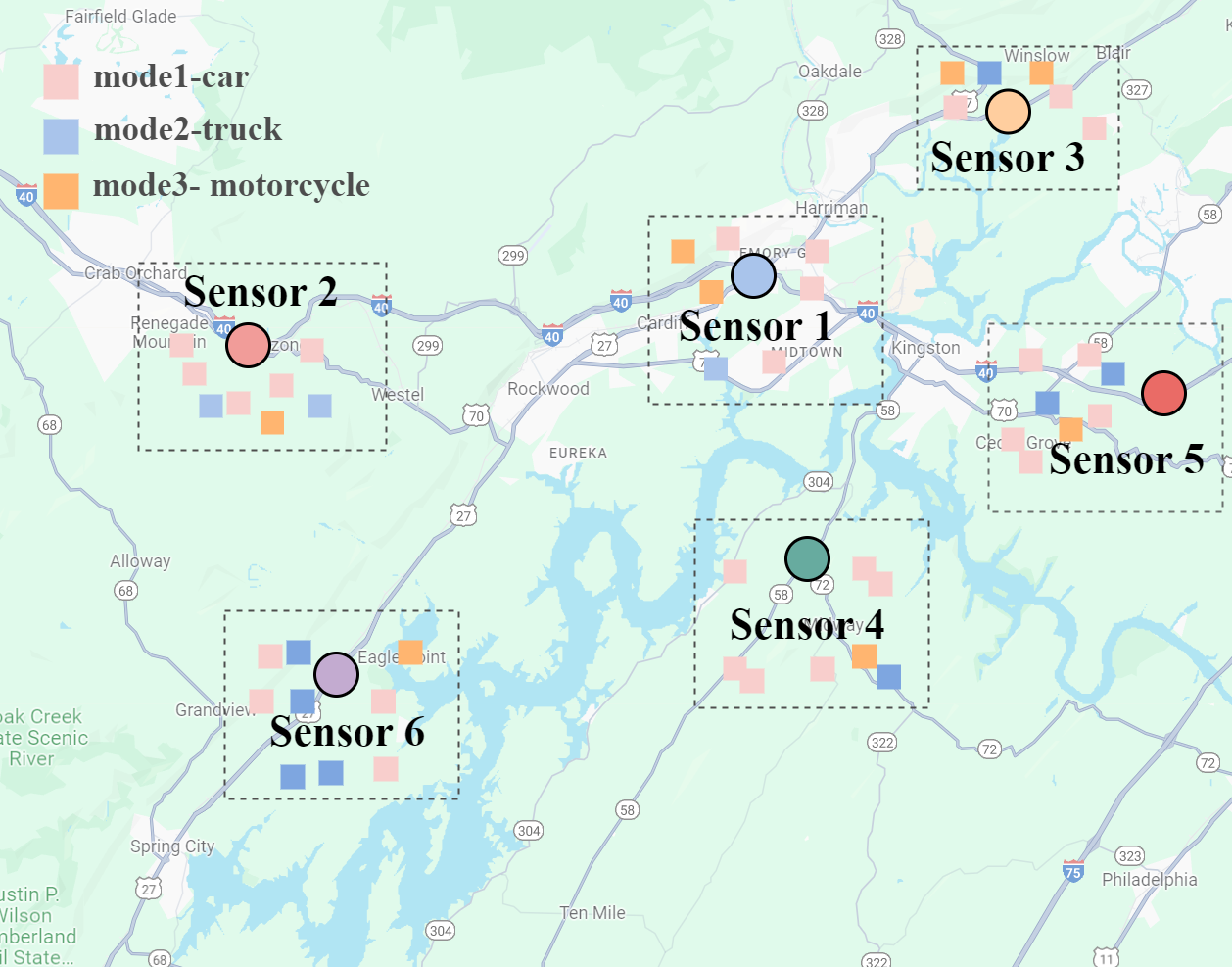}
	\end{subfigure}%
	\begin{subfigure}{0.24\textwidth}
		\centering
		\includegraphics[width=\linewidth]{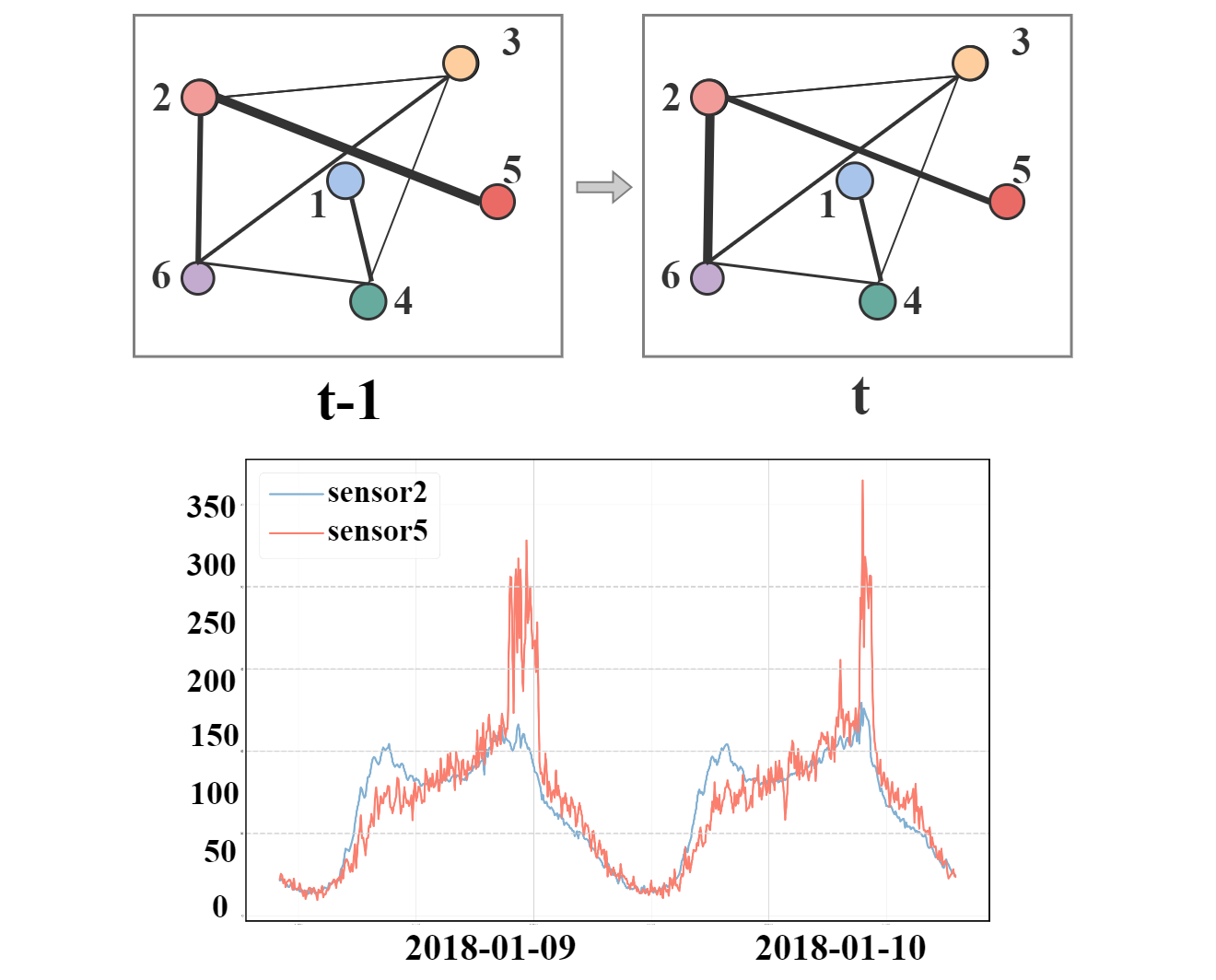}
	\end{subfigure}
	\caption{An example of dynamic spatio-temporal relationships in multimodal transportation}
 \label{figure1}
\end{figure}

Figure \ref{figure1} illustrates a typical traffic system with multiple coexisting transportation patterns and its dynamic spatio-temporal relationships, strategically deploying traffic sensors across the road network to record traffic flow data. These data represent the count of vehicles crossing specific spatial regions within defined time intervals. At time t-1, different nodes exhibit distinct distributions of traffic patterns. Sensors 2 and 5 share similar traffic pattern compositions, despite being relatively distant spatially. There remains a strong correlation in traffic flow between Sensors 2 and 5 at time t-1. This correlation is dynamic; compared to time step t-1, the correlation between sensors 2 and 5 weakens at time step t, while the correlation between sensors 2 and 6 strengthens. Therefore, passive integration of spatial and temporal dimensions is needed to consider how to capture the dynamic adaptation relationships of different traffic pattern flows at nodes.

However, recent research often entails constructing a spatial graph with a predefined or data-adaptive adjacency matrix\cite{bigDataTraffic,ST-GRAT, spatialTemporalGraph}. Graph neural networks are utilized to capture the correlations between spatial nodes at each time step, and recurrent neural networks (RNN) or Transformers are employed to establish connections between nodes and itself in adjacent time steps, thereby capturing information from temporally adjacent nodes\cite{ST-MetaNet,STGCN,AGCRN}. Nevertheless, these methods have not sufficiently extracted the spatio-temporal relationships between different time intervals and nodes, including the dynamic variations in spatio-temporal correlations\cite{CNN,CNN2,GCN}. Therefore, we posit that more effective methods are requisite to extract spatio-temporal relationships between different time intervals and nodes, accounting for dynamic changes in spatio-temporal relationships while considering the allocation of distinct spatio-temporal fusion graphs for decoupled traffic patterns.

We devise a Traffic Flow Decoupling Spatio-temporal Fusion Graph to effectively model the dynamic spatio-temporal implicit relationships in a multimodal transportation system.  This graph decouples the traffic flow into distinct traffic pattern streams through node embeddings and time embeddings. Each traffic pattern stream is then assigned a unique adaptive spatio-temporal fusion graph for residual graph convolution. This approach facilitates capturing dynamic implicit relationships between time and space. Subsequently, the outputs from multiple components are temporally connected, with a Recurrent Neural Network (RNN) capturing short-term temporal information. This information is concatenated with periodic temporal features from time embeddings to capture multiscale features. 

In summary, our main contributions are as follows:
\begin{itemize}
\item We propose a novel method to improve traffic flow prediction through adaptive spatiotemporal fusion graphs. This method captures spatiotemporal characteristics by constructing time and space feature matrices and their dynamic fusion, using a cross-attention mechanism to adjust the dynamic focus, allowing the graph to adapt to different traffic patterns.

\item We introduce a traffic pattern decoupling module that utilizes spatiotemporal features to decompose complex traffic flows into independent representations for each pattern. In the residual graph convolution module, the model captures spatiotemporal features with pattern labels from different traffic flows within the generated dynamic spatiotemporal graph.

\item Extensive research on four large-scale real-world datasets shows that our framework consistently outperforms existing methods in terms of performance.
\end{itemize}
\begin{figure*}
    \centering
    \includegraphics[width=.9\textwidth]{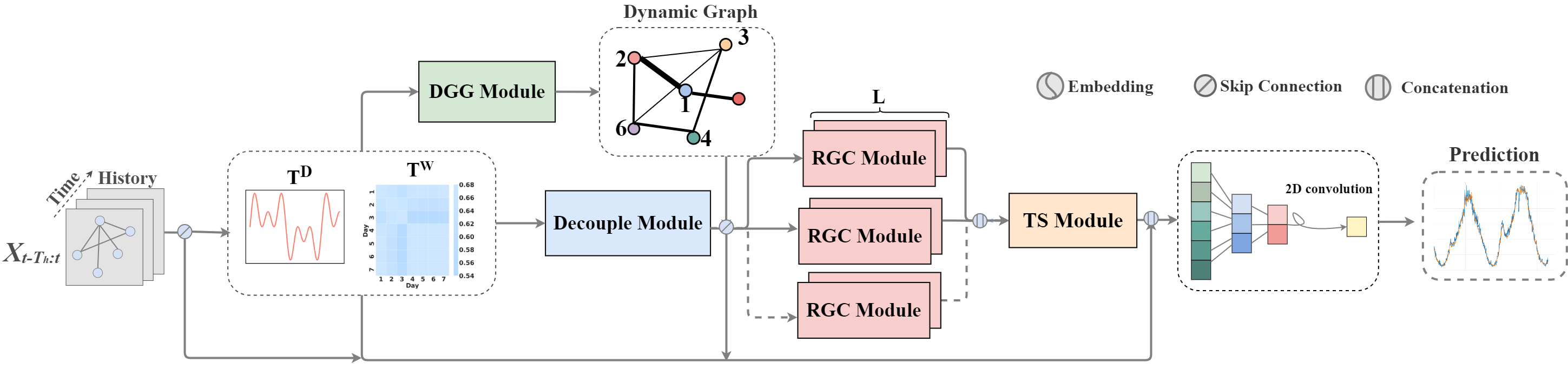}
    \caption{The framework of SFADNet}
    \label{figure2}
\end{figure*}
\section{Methodology}
\subsection{Dynamic Graph Generation Module}
In Figure \ref{figure2}, we showcase the overall architecture of our model. We utilize an efficient approach to generate spatial graphs by calculating pairwise relationships exclusively among selected subsets of nodes. The formula is detailed below:

\begin{equation}
    \begin{gathered}
        \mathbf{M}_1 = tanh(\alpha \mathbf{F}_1\mathbf{W}_1) \\
        \mathbf{M}_2 = tanh(\alpha\mathbf{F}_2\mathbf{W}_2) \\ 
        \mathbf{A} = ReLU(tanh(\alpha(\mathbf{M}_1\mathbf{M}_2^T-\mathbf{M}_1^T\mathbf{M}_2))) \\
        \hat{\mathbf{A}} = \left\{
        \begin{aligned}
            &\mathbf{A}^s_{ij},j \in argtopk(\mathbf{A}^s[i,:])\\
            &0,otherwise
        \end{aligned}
        \right.
    \end{gathered}
\end{equation}

 Matrices $\mathbf{F}_1 \in \mathbb{R}^{N \times N_f}$ and $\mathbf{F}_2 \in \mathbb{R}^{N \times N_f}$ represent the spatial or temporal feature embedding matrices of nodes, while $\mathbf{W_1}$ and $\mathbf{W_2}$ denote the parameter matrices of fully connected layers. By constructing the asymmetric matrix $\mathbf{\hat{A}}$ using the subtraction terms $\mathbf{M_1}$ and $\mathbf{M_2}$ along with an activation function, we capture the similarity differences, indicating the unidirectional influence relationships of temporal or spatial features between nodes. Then, we only retain the top K elements of each row in the adjacency matrix $\mathbf{\hat{A}}$ to sparsify the matrix, thereby reducing computational complexity.

For the spatial feature embedding matrices, we use randomly initialized node embeddings $\mathbf{E}_1 \in \mathbb{R}^{N \times N_d}$ and $\mathbf{E}_2 \in \mathbb{R}^{N \times N_d}$. The construction of the temporal feature matrix is as follows:

\begin{equation}
    \begin{gathered}
        \mathbf{T} = \frac{1}{T_h}\sum_{t=1}^{\mathbf{T}_h}\mathbf{T}_t 
    \end{gathered}
\end{equation}

$\mathbf{T_t}$ represents the daily embedding matrix $\mathbf{T_t^D} \in \mathbb{R}^{T_h \times N \times D}$ or the weekly embedding matrix $\mathbf{T_t^W} \in \mathbb{R}^{T_h \times N \times D}$ obtained from the time pool based on daily and weekly indices. It is then averaged along the time dimension $T_h$ to obtain a continuous real-valued vector that represents periodic features.

\begin{equation}
\begin{gathered}
        \text{Multihead}(\hat{\mathbf{A}}^s,\hat{\mathbf{A}}^t,\hat{\mathbf{A}}^f) =(\text{head}_1||\text{head}_2\ldots ||\text{head}_s)W_O  \\
        \text{head}_s = \text{softmax}(\frac{(\hat{\mathbf{A}}^s\mathbf{W}_Q^s)(\hat{\mathbf{A}}^t\mathbf{W}_K^s)}{\sqrt{d}}(\hat{\mathbf{A}}^f\mathbf{W}_V^s) \\
    \end{gathered}
\end{equation}
To perform weighted fusion on the asymmetric spatiotemporal feature matrices, we introduce multi-head attention to learn in parallel across different subspaces. Here, \(\mathbf{\hat{A}}^s\) and \(\mathbf{\hat{A}}^t\) represent the temporal and spatial feature matrices, respectively. The fused feature matrix \(\mathbf{A}^f\) is obtained using the formula: $\mathbf{\hat{A}}^f = \text{ReLU}(\tanh(\beta \hat{\mathbf{A}}^s \hat{\mathbf{A}}^t))$
to enhance the spatiotemporal feature representation and improve the robustness of the model.
\subsection{Decouple Module}
The decoupling module used to classify traffic flow into $\textbf{M}$ types of traffic patterns can be expressed as follows:
\begin{equation}
\begin{gathered}
\mathbf{\Omega}_{t,i} = \text{Sigmoid}((\text{ReLU}(\mathbf{\mathbf{T}}_{(t,i)}^D||\mathbf{\mathbf{T}}_{(t,i)}^W||\mathbf{E}_i)\mathbf{W}_1)\mathbf{W}_2) \\
\mathbf{\mathcal{X}}_g = \mathbf{\mathcal{X}}_{t-T_h:t}\odot\mathbf{\Omega}, g\in \{1,2,\ldots, G-1\} \\
\mathbf{\mathcal{X}}_{G} = \mathbf{\mathcal{X}}_{t-T_h:t}-\displaystyle\sum_{g=1}^{G-1}\mathbf{\mathcal{X}}_g 
\end{gathered}
\end{equation}

The matrix $\Omega \in \mathbb{R}^{T_h \times N \times 1}$ represents the ratio of a specific traffic pattern relative to the total traffic flow $\mathcal{X}_{t-T_h:t}$. By element-wise multiplying the decoupling ratio matrix $\Omega$ with the original traffic flow, we obtain the traffic flow proportion of the $g$-th traffic pattern $\mathcal{X}_g \in \mathbb{R}^{T_h \times N \times C}$.

\subsection{Residual Graph Convolution Module}
\begin{figure}[h]
	\centering
	\begin{subfigure}{0.25\textwidth}
		\centering
		\includegraphics[width=\linewidth]{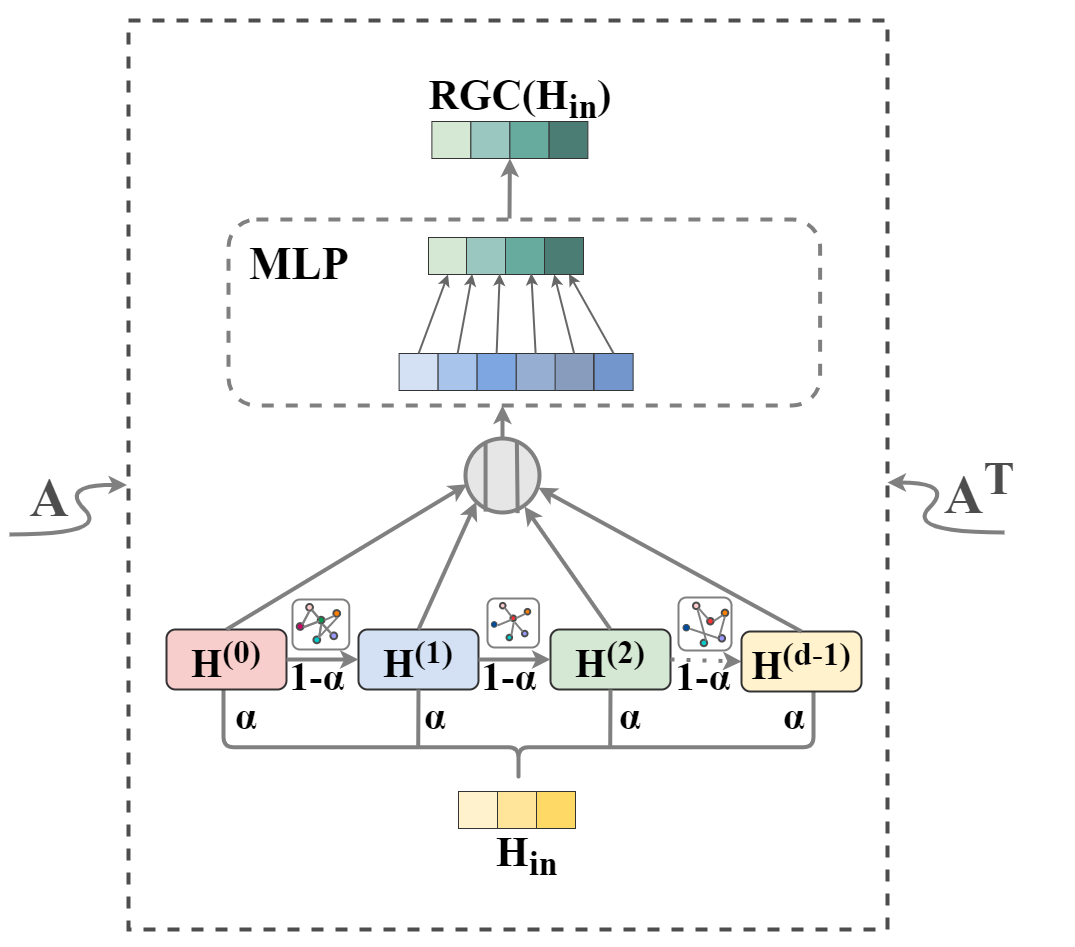}
		\caption{RGC Module}
		\label{figure3a}
	\end{subfigure}%
	\begin{subfigure}{0.185\textwidth}
		\centering
		\includegraphics[width=\linewidth]{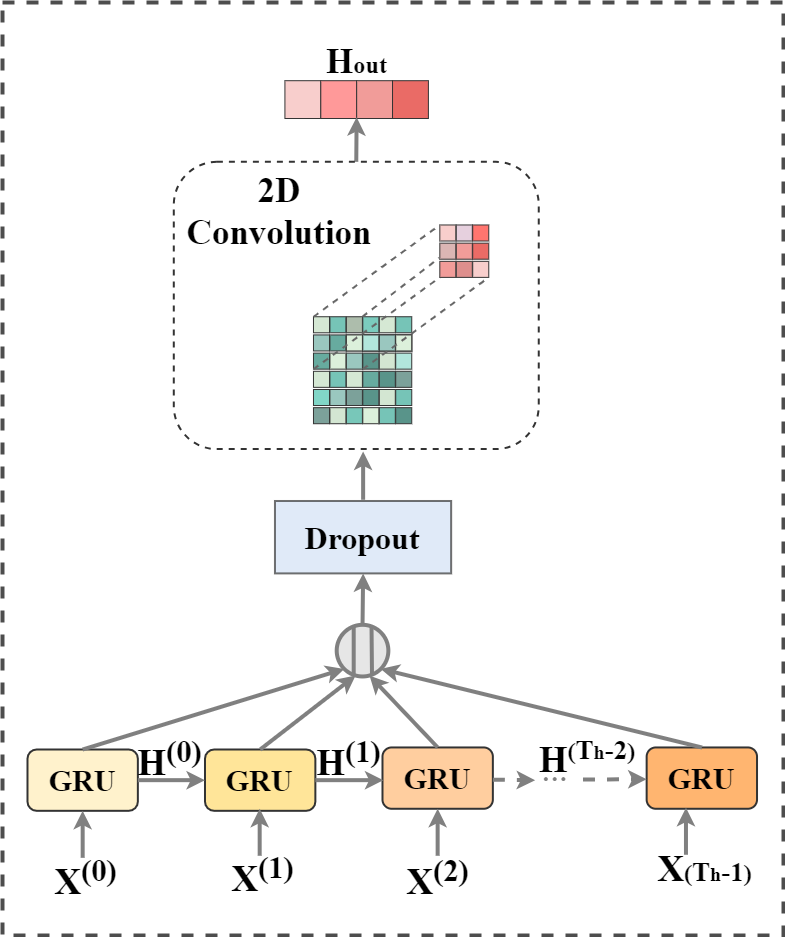}
		\caption{TS Module}
		\label{figure3b}
	\end{subfigure}
	\caption{Diagram of Model Components}
	\label{figure3}
\end{figure}

Figure \ref{figure3a} illustrates the RGC Module, intended for processing various traffic pattern flows using the adaptive spatio-temporal fusion graph $\mathbf{A}$ generated in the DGG module.

\begin{equation}
    \begin{gathered}
    \widetilde{\mathbf{A}} = \mathbf{A} + \mathbf{I} \\
\mathbf{H}^{(k)}_G={\mathrm{\gamma}} \mathbf{H}_{\text{in}} + (1-\mathrm{\gamma})\left(\widetilde{\mathbf{D}}^{-1} \widetilde{\mathbf{A}}\right) \mathbf{H}^{(k-1)}_G \\
RGC(\mathbf{H}_{in}) = (\mathbf{H}_G^{(0)} \, \Vert \, \mathbf{H}_G^{(1)} \, \Vert \, \ldots \, \Vert \, \mathbf{H}_G^{(K-1)})\mathbf{W}
\\
    \end{gathered}
\end{equation}
In the formula, $\widetilde{A} \in \mathbb{R}^{N \times N}$ is the adjacency matrix with self-connections, and the degree matrix $\widetilde{D} \in \mathbb{R}^{N \times 1}$ satisfies $\widetilde{D}_{ii} = \sum_j \widetilde{A}_{ij}$. $k$ represents the propagation depth and the parameter $\gamma$ indicates the proportion of the input node $\mathbf{H}_{\text{in}}$ to be retained during node updates. The purpose of this step is to find a balance in node representation updates, ensuring the preservation of both local features and the facilitation of deep exploration.

\subsection{Temporal Sequence Module}
The spatial-temporal features extracted after the individual traffic flow data has undergone the residual graph convolution module are concatenated. The formula is as follows:
\begin{equation}
    \begin{gathered}
     \hat{\mathbf{\mathcal{X}}_i} = \underbrace{\{RGC(\mathbf{\mathcal{X}}_i)||RGC(\mathbf{\mathcal{X}}_i)\ldots||RGC(\mathbf{\mathcal{X}}_i)\}}_{\mathbf{M}}\\
        \mathbf{\mathcal{X}}_{out} = (\hat{\mathbf{\mathcal{X}}_1}||\hat{\mathbf{\mathcal{X}}_2}\ldots ||\hat{\mathbf{\mathcal{X}}_G})\\
    \end{gathered}
\end{equation}
Where $M$ is the number of iterations in the residual graph convolution module, and $G$ is the number of decoupled traffic patterns, we obtain the concatenated result $\mathcal{X}_{\text{out}} \in \mathbb{R}^{T_h \times N \times (M\cdot d)}$ from multiple iterations of each traffic pattern in the Residual Graph Convolution (RGC) module. This concatenated result serves as the input to the Temporal Sequence (TS) module, as depicted in Figure \ref{figure3b}. The GRU in the temporal sequence module employs gating mechanisms, allowing selective storage and updating of hidden states, facilitating the capture of periodic patterns and dependencies in the input traffic flow data.
\begin{equation}
\begin{gathered}
z_t = \text{sigmoid}(\mathbf{W}_z \mathcal{X}_{out}^{(t)} + \mathbf{U}_z \mathbf{H}^{(t-1)} + b_z) \\
r_t = \text{sigmoid}(\mathbf{W}_r\mathcal{X}_{out}^{(t)} + \mathbf{U}_r\mathbf{H}^{(t-1)} + b_r) \\
\mathbf{\tilde{H}}_t = \text{tanh}(\mathbf{W}_h \mathcal{X}_{out}^{(t)} + \mathbf{U}_h (r_t\odot \mathbf{H}^{(t-1)}) + b_h) \\
\mathbf{H}^{(t)} = (1 - z_t) \odot \mathbf{H}^{(t-1)} + z_t \odot \mathbf{\tilde{H}}^{(t)} \\
\end{gathered}
\end{equation}
Where $\mathcal{X}_{\text{out}}^{(t)} \in \mathbb{R}^{N \times (M\cdot d)}$ represents the traffic flow input at each time step. We obtain the current time step output $\mathbf{H}^{(t)} \in \mathbb{R}^{N \times (M\cdot d)}$ by updating the update gate $z_t$ and and the reset gate $r_t$, and stacking the outputs of $T_h$  time steps together to form the output tensor $\mathbf{H}_{\text{out}} \in \mathbb{R}^{T_h \times N \times (M\cdot d)}$:
\begin{equation}
    \begin{gathered}
        \mathbf{H}_{out} = \text{dropout}([\mathbf{H}^{(0)}\mathbf{H}^{(1)}\mathbf{H}^{(2)}\ldots \mathbf{H}^{(T_h-1)}])
    \end{gathered}
\end{equation}
\subsection{Output and Training Strategy}
We establish skip connections for the crucial information flow within the model.
\begin{equation}
    \begin{gathered}
        \mathbf{H} = \{\mathbf{H}_{out}||\mathcal{X}_{out}||\mathcal{X}_{t-T_h:t}||\mathbf{T}^D_t||\mathbf{T}^W_t\}
    \end{gathered}
\label{equation11}
\end{equation}
After obtaining the tensor $\mathbf{H}$ through skip connections, a regression layer is formed by applying both fully connected and two-dimensional convolutional operations, resulting in the predicted state $\mathcal{X}_{t:t+T_f} \in \mathbb{R}^{T_f \times N \times C}$:
\begin{equation}
    \begin{gathered}
        \mathcal{X}_{t:t+T_f} = [(ReLU(ReLU(\mathbf{H})\mathbf{W}_1)\mathbf{W}_2)\ast \mathbf{W}]^T\\
    \end{gathered}
\end{equation}

We adopt curriculum learning\cite{cl}, a general and effective training strategy, to train the proposed model, and use warm-up steps\cite{resnet} to help the model converge better, employing gradient descent to minimize the loss function and optimize model parameters.

\section{Experiments}
\subsection{Datasets}
We utilize four datasets from the California Department of Transportation's (CalTrans) Performance Measurement System (PEMS), namely PEMS03, PEMS04, PEMS07, and PEMS08. The original data was aggregated at 5-minute intervals, and a predefined spatial graph was constructed based on the actual road network, as detailed in Table \ref{table1}. Our objective involved predicting future traffic flow for the next 60 minutes (12 time steps) based on the preceding 60 minutes' (12 time steps) historical traffic flow. 
\begin{table*}[h]
  \centering
  \caption{Performance Comparison Against Baselines}
  \resizebox{.95\textwidth}{!}{
    \begin{tabular}{ccccccccccccc}
      \toprule[1.5pt]
      \multirow{2}{*}{Models} & \multicolumn{3}{c}{PEMS03} & \multicolumn{3}{c}{PEMS04} & \multicolumn{3}{c}{PEMS07} & \multicolumn{3}{c}{PEMS08}\\
      \cmidrule[0.75pt]{2-13}
      & MAE & RMSE & MAPE(\%) & MAE & RMSE & MAPE(\%) & MAE & RMSE & MAPE(\%) & MAE & RMSE & MAPE(\%)\\
      \midrule[0.75pt]
      FC-LSTM &21.33 &35.11 &23.33 &27.14 &41.59 &18.20 &29.98 &45.94 &13.20 &22.20 &34.06 &14.20 \\
      DSANet &21.29 &34.55 &23.21 &22.79 &35.77 &16.03 &31.36 &49.11 &14.43 &17.14 &26.96 &11.32 \\
      GraphWaveNet &19.12 &32.77 &18.89 &24.89 &39.66 &17.29 &26.39 &41.50 &11.97 &18.28 &30.05 &12.15  \\
      DCRNN  &18.30 &29.74 &17.86 &23.54 &36.25 &17.18 &23.87  &37.27 &10.50 &18.41 &28.28 &12.17 \\
      ASTGCN &17.34 &29.56 &17.21 &22.92 &35.22 &16.56 &24.01 &37.87 &10.73 &18.25 &28.06 &11.64 \\
      STFGNN &16.77 &28.34 &16.30 &19.83 &31.88 &13.02 &22.07 &35.80 &9.21  
      &16.64 &26.22 &10.60 \\
      STGODE &16.50 &27.84 &16.69 &20.84 &32.82 &13.77 &22.59 &37.54 &10.14
      &16.81 &25.97 &10.62 \\
      STG-NCDE &15.57&27.09&15.06&19.21&31.09&12.76&20.53&33.84&8.80
      &15.54&24.81&9.92 \\
      DSTAGNN &15.57&27.21&\textbf{14.68}&19.30&31.46&12.70&21.42&34.51&9.01
      &15.67&24.77&9.94 \\ST-AE&15.44&26.29&15.20&19.74&31.14&15.40&22.75&35.56&10.18&15.96&25.03&11.34 \\
      \midrule[0.75pt]
      \textbf{SFADNet} &\textbf{14.66} &\textbf{24.50} &15.04 &\textbf{18.32} &\textbf{30.19} &\textbf{12.21} &\textbf{19.44} &\textbf{32.77} &\textbf{8.20}
      &\textbf{13.65} &\textbf{23.23} &\textbf{8.98} \\
      \bottomrule[1.5pt]
    \end{tabular}
  }
  \label{table3}
\end{table*}

\begin{table}[h]
  \centering
  \caption{Datasets description.}
  \resizebox{.45\textwidth}{!}{
  \begin{tabular}{ccccc}
    \toprule[1.5pt] 
    Dataset &\#Nodes &\#Edges &\#Days &\#Data\\
        \midrule[0.75pt] 
    PEMS03 &358 &547 &91 &26208 \\
    PEMS04 &307 &340 &59 &16992 \\
    PEMS07 &883 &866 &98 &28224 \\
    PEMS08 &170 &295 &62 &17856 \\
    \bottomrule[1.5pt] 
  \end{tabular}
  }
  \label{table1}
\end{table}

\subsection{Baseline Methods}
SFADNet's performance is compared against the following baselines, including both traditional models and state-of-the-art approaches: FC-LSTM\cite{FC-LSTM}, DSANet\cite{DSANet}, GraphWaveNet\cite{GraphWaveNet}, DCRNN\cite{DCRNN}, ASTGCN\cite{ASTGCN}, STFGNN\cite{STFGNN}, STGODE\cite{STGODE}, STG-NCDE\cite{STG-NCDE}, DSTAGNN\cite{DSTAGNN}, ST-AE\cite{ST-AE}
\subsection{Experiment Settings}
The experiments are conducted on a system equipped with an NVIDIA GeForce RTX 4090 GPU. The datasets PEMS03, PEMS04, PEMS07, and PEMS08 are divided into training, validation, and test sets in a ratio of 6:2:2. The models are trained using the Adam optimization algorithm with a weight decay (wdecay) of \(1.0 \times 10^{-5}\) and an epsilon (eps) of \(1.0 \times 10^{-8}\). Additionally, a learning rate scheduler is employed to automatically adjust the learning rate with a decay rate of 0.5. Implement a warm-up strategy during the first 20 training cycles, followed by curriculum learning with a length of 3. The configurations of model-related parameters for the four datasets are detailed in Table \ref{table2}. The source code of SFADNet is  available: https://github.com/meiwu5/SFADNet.git

\begin{table}[h]
    \centering
    \caption{Model Configuration for Each Dataset}
      \resizebox{.48\textwidth}{!}{
    \begin{tabular}{cccccccc}
        \toprule[1.5pt]
        Dataset & Batch Size & Learning Rate & $N_d$ & D & $K_s$ & $K_t$ & G \\
        \midrule
        PEMS03 & 32 & 0.004 & 12 & 12 & 10 & 10 & 2 \\
        PEMS04 & 32 & 0.004 & 12 & 12 & 10 & 10 & 2 \\
        PEMS07 & 12 & 0.004 & 12 & 12 & 10 & 10 & 2 \\
        PEMS08 & 32 & 0.004 & 10 & 10 & 10 & 10 & 2 \\
        \bottomrule[1.5pt]
    \end{tabular}
    }
    \label{table2}
\end{table}
\subsection{Experiment Results}
Table \ref{table3} presents a comparison of different methods on MAE, RMSE, and MAPE across four datasets. Compared to non-graph models like FC-LSTM and DSANet, as well as predefined graph-based models like DCRNN, methods such as STGODE and STG-NCDE that introduce ODE and NCDE to expand their spatial receptive fields, attention-based models like ST-AE, and dynamically aware graph-based models like DSTAGNN, our SFADNet achieves the best results in all metrics across the four datasets.

\subsection{Ablation Experiments}

\begin{table}[h]
  \centering
  \caption{Spatiotemporal Graph Ablation Experiment}
  \resizebox{.45\textwidth}{!}{
    \begin{tabular}{ccccccc}
      \toprule[1.5pt]
      \multirow{2}{*}{Models} & \multicolumn{3}{c}{PEMS03} & \multicolumn{3}{c}{PEMS08} \\
      \cmidrule[0.75pt]{2-4} \cmidrule[0.75pt]{5-7}
      & MAE & RMSE & MAPE(\%) & MAE & RMSE & MAPE(\%) \\
      \midrule[0.75pt]
      use pg &15.22 &26.67 &15.91 &14.14 &23.46 &9.17 \\
      use tg &15.89 &27.22 &16.14 &13.99 &23.46 &9.16 \\
      use sg &15.20 &26.16 &15.61 &13.97 &23.12 &9.35 \\
      SFADNet &14.66 &24.50 &15.04 &13.65 &23.23 &8.98 \\
      \bottomrule[1.5pt]
    \end{tabular}
  }
  \label{table4}
\end{table}

\begin{itemize}
    \item \textbf{use pg}: This variant utilizes predefined graphs.
    \item \textbf{use tg}: This variant generates time graphs using only the time feature matrix.
    \item \textbf{use sg}: This variant generates a spatial graph using only the spatial feature matrix.
\end{itemize}

\begin{table}[h]
  \centering
  \caption{Decoupling Module Ablation Experiment}
  \resizebox{.45\textwidth}{!}{
    \begin{tabular}{ccccccc}
      \toprule[1.5pt]
      \multirow{2}{*}{Models} & \multicolumn{3}{c}{PEMS04} & \multicolumn{3}{c}{PEMS08} \\
      \cmidrule[0.75pt]{2-4} \cmidrule[0.75pt]{5-7}
      & MAE & RMSE & MAPE(\%) & MAE & RMSE & MAPE(\%) \\
      \midrule[0.75pt]
      w/o DM & 18.75&30.41&12.79 &14.68 &23.97 &9.58 \\
      $\textbf{SFADNet}_{M=2}$ &18.46 &30.23 &12.27 &13.65 &23.23 &8.98 \\
      $\textbf{SFADNet}_{M=3}$ & 18.32 &30.19 &12.21 &13.73 &23.26 &9.01 \\
      \bottomrule[1.5pt]
    \end{tabular}
  }
  \label{table5}
\end{table}

\begin{itemize}
    \item \textbf{w/o DM}: This variant does not use the Decouple Module.
    \item \textbf{$\textbf{SFADNet}_{M=2}$}: This variant decouples into two traffic patterns.
    \item \textbf{$\textbf{SFADNet}_{M=3}$}: This variant decouples into three traffic patterns.
\end{itemize}

\section{Conclusion}
This paper introduces SFADNet, a spatiotemporal fusion graph decoupling network that incorporates attention mechanisms. SFADNet constructs an adaptive spatiotemporal fusion graph and classifies traffic flow into different patterns based on temporal flow and spatial node attributes. It employs a traffic pattern adaptive graph that combines temporal and spatial graphs for residual graph convolution, effectively capturing residual information through a temporal sequence module.

\section*{ACKNOWLEDGMENT}
This work is supported by the Industry- University Cooperative Education Project of Ministry of Education of China under Grant 231106093150720 and 220903242265640.


\begin{thebibliography}{00}
\bibitem{trafficIncidentDetection} X. Han, T. Grubenmann, R. Cheng, S. C. Wong, X. Li, and W. Sun, ``Traffic Incident Detection: A Trajectory-based Approach,'' in \textit{Proceedings of the 2020 IEEE 36th International Conference on Data Engineering (ICDE)}, pp. 1866--1869, 2020. [DOI: 10.1109/ICDE48307.2020.00190].

\bibitem{CABIN} T. Qian, F. Wang, Y. Xu, Y. Jiang, T. Sun, and Y. Yu, ``CABIN: A Novel Cooperative Attention Based Location Prediction Network Using Internal-External Trajectory Dependencies,'' in \textit{Artificial Neural Networks and Machine Learning -- ICANN 2020}, pp. 521--532, 2020. [ISBN: 978-3-030-61616-8].

\bibitem{DeepTRANS} L. Tran, M. Y. Mun, M. Lim, J. Yamato, N. Huh, and C. Shahabi, ``DeepTRANS: A Deep Learning System for Public Bus Travel Time Estimation Using Traffic Forecasting,'' \textit{Proc. VLDB Endow.}, vol. 13, no. 12, pp. 2957--2960, Aug. 2020. [DOI: 10.14778/3415478.3415518].

\bibitem{jointPredictionModel} H. Yuan, G. Li, Z. Bao, and L. Feng, ``An Effective Joint Prediction Model for Travel Demands and Traffic Flows,'' in \textit{Proceedings of the 2021 IEEE 37th International Conference on Data Engineering (ICDE)}, pp. 348--359, 2021. [DOI: 10.1109/ICDE51399.2021.00037].

\bibitem{smoothingModel} G. Kitagawa and W. Gersch, ``A Smoothness Priors–State Space Modeling of Time Series with Trend and Seasonality,'' \textit{Journal of the American Statistical Association}, vol. 79, no. 386, pp. 378--389, 1984. [DOI: 10.1080/01621459.1984.10478060]..

\bibitem{deepLearningExperiment} H. Lee, C. Park, S. Jin, H. Chu, J. Choo, and S. Ko, ``An Empirical Experiment on Deep Learning Models for Predicting Traffic Data,'' in \textit{Proceedings of the 2021 IEEE 37th International Conference on Data Engineering (ICDE)}, pp. 1817--1822, 2021. [DOI: 10.1109/ICDE51399.2021.00160].

\bibitem{bigDataTraffic} Y. Lv, Y. Duan, W. Kang, Z. Li, and F.-Y. Wang, ``Traffic Flow Prediction With Big Data: A Deep Learning Approach,'' \textit{IEEE Transactions on Intelligent Transportation Systems}, vol. 16, no. 2, pp. 865--873, 2015. [DOI: 10.1109/TITS.2014.2345663].

\bibitem{ST-GRAT} C. Park, C. Lee, H. Bahng, Y. Tae, S. Jin, K. Kim, J. Ko, and J. Choo, ``ST-GRAT: A Novel Spatio-temporal Graph Attention Networks for Accurately Forecasting Dynamically Changing Road Speed,'' in \textit{Proceedings of the 29th ACM International Conference on Information \& Knowledge Management}, pp. 1215--1224, 2020. [DOI: 10.1145/3340531.3411940].

\bibitem{spatialTemporalGraph} S. Guo, Y. Lin, H. Wan, X. Li, and G. Cong, ``Learning Dynamics and Heterogeneity of Spatial-Temporal Graph Data for Traffic Forecasting,'' \textit{IEEE Transactions on Knowledge and Data Engineering}, vol. 34, no. 11, pp. 5415--5428, 2022. [DOI: 10.1109/TKDE.2021.3056502].

\bibitem{ST-MetaNet} Z. Pan, Y. Liang, W. Wang, Y. Yu, Y. Zheng, and J. Zhang, ``Urban Traffic Prediction from Spatio-Temporal Data Using Deep Meta Learning,'' in \textit{Proceedings of the 25th ACM SIGKDD International Conference on Knowledge Discovery \& Data Mining}, pp. 1720--1730, 2019. [DOI: 10.1145/3292500.3330884].

\bibitem{STGCN} B. Yu, H. Yin, and Z. Zhu, ``Spatio-Temporal Graph Convolutional Networks: A Deep Learning Framework for Traffic Forecasting,'' in \textit{Proceedings of the Twenty-Seventh International Joint Conference on Artificial Intelligence}, 2018. [DOI: 10.24963/ijcai.2018/505].

\bibitem{AGCRN} L. Bai, L. Yao, C. Li, X. Wang, and C. Wang, ``Adaptive Graph Convolutional Recurrent Network for Traffic Forecasting,'' \textit{Advances in Neural Information Processing Systems}, vol. 33, pp. 17804--17815, 2020.



\bibitem{CNN} W. Zhang, Y. Yu, Y. Qi, F. Shu, and Y. Wang, ``Short-term traffic flow prediction based on spatio-temporal analysis and CNN deep learning,'' in \textit{Transportmetrica A: Transport Science}, vol. 15, no. 2, pp. 1688--1711, 2019. [DOI: 10.1080/23249935.2019.1637966].

\bibitem{CNN2} M. Cao, V. O. K. Li, and V. W. S. Chan, ``A CNN-LSTM Model for Traffic Speed Prediction,'' in \textit{2020 IEEE 91st Vehicular Technology Conference (VTC2020-Spring)}, pp. 1--5, 2020. [DOI: 10.1109/VTC2020-Spring48590.2020.9129440].

\bibitem{GCN} B. Yu, Y. Lee, and K. Sohn, ``Forecasting road traffic speeds by considering area-wide spatio-temporal dependencies based on a graph convolutional neural network (GCN),'' in \textit{Transportation Research Part C: Emerging Technologies}, vol. 114, pp. 189--204, 2020. [DOI: 10.1016/j.trc.2020.02.013].

\bibitem{FC-LSTM} I. Sutskever, O. Vinyals, and Q. V. Le, ``Sequence to Sequence Learning with Neural Networks,'' in \textit{Advances in Neural Information Processing Systems}, vol. 27, Z. Ghahramani, M. Welling, C. Cortes, N. Lawrence, and K.Q. Weinberger, Eds. Curran Associates, Inc., 2014.

\bibitem{DSANet} S. Huang, D. Wang, X. Wu, and A. Tang, ``DSANet: Dual Self-Attention Network for Multivariate Time Series Forecasting,'' in \textit{Proceedings of the 28th ACM International Conference on Information and Knowledge Management}, Beijing, China, 2019, pp. 2129-2132.

\bibitem{GraphWaveNet} Z. Wu, S. Pan, G. Long, J. Jiang, and C. Zhang, ``Graph Wavenet for Deep Spatial-Temporal Graph Modeling,'' in \textit{Proceedings of the 28th International Joint Conference on Artificial Intelligence}, 2019, pp. 1907-1913.

\bibitem{DCRNN} Y. Li, R. Yu, C. Shahabi, and Y. Liu, ``Diffusion Convolutional Recurrent Neural Network: Data-Driven Traffic Forecasting,'' in \textit{International Conference on Learning Representations}, 2018.

\bibitem{ASTGCN} S. Guo, Y. Lin, N. Feng, C. Song, and H. Wan, ``Attention Based Spatial-Temporal Graph Convolutional Networks for Traffic Flow Forecasting,'' \textit{Proceedings of the AAAI Conference on Artificial Intelligence}, vol. 33, no. 01, pp. 922-929, July 2019.

\bibitem{STFGNN} M. Li and Z. Zhu, ``Spatial-Temporal Fusion Graph Neural Networks for Traffic Flow Forecasting,'' \textit{Proceedings of the AAAI Conference on Artificial Intelligence}, vol. 35, no. 5, pp. 4189-4196, May 2021.

\bibitem{STGODE} Z. Fang, Q. Long, G. Song, and K. Xie, ``Spatial-Temporal Graph ODE Networks for Traffic Flow Forecasting,'' in \textit{Proceedings of the 27th ACM SIGKDD Conference on Knowledge Discovery \& Data Mining}, New York, NY, USA, 2021, pp. 364-373.

\bibitem{STG-NCDE} J. Choi, H. Choi, J. Hwang, and N. Park, ``Graph Neural Controlled Differential Equations for Traffic Forecasting,'' \textit{Proceedings of the AAAI Conference on Artificial Intelligence}, vol. 36, no. 6, pp. 6367-6374, June 2022.

\bibitem{DSTAGNN} S. Lan, Y. Ma, W. Huang, W. Wang, H. Yang, and P. Li, ``DSTAGNN: Dynamic Spatial-Temporal Aware Graph Neural Network for Traffic Flow Forecasting,'' in \textit{Proceedings of the 39th International Conference on Machine Learning}, vol. 162, pp. 11906--11917, 2022.

\bibitem{ST-AE} M. Liu, T. Zhu, J. Ye, Q. Meng, L. Sun, and B. Du, ``Spatio-Temporal AutoEncoder for Traffic Flow Prediction,'' \textit{IEEE Transactions on Intelligent Transportation Systems}, vol. 24, no. 5, pp. 5516-5526, 2023. [DOI: 10.1109/TITS.2023.3243913].

\bibitem{cl} Z. Wu, S. Pan, G. Long, J. Jiang, X. Chang, and C. Zhang, ``Connecting the Dots: Multivariate Time Series Forecasting with Graph Neural Networks,'' in \textit{Proceedings of the 26th ACM SIGKDD International Conference on Knowledge Discovery \& Data Mining}, pp. 753--763, 2020. [DOI: 10.1145/3394486.3403118].

\bibitem{resnet} K. He, X. Zhang, S. Ren, and J. Sun, ``Deep Residual Learning for Image Recognition,'' in \textit{Proceedings of the IEEE Conference on Computer Vision and Pattern Recognition (CVPR)}, June 2016.

\end{thebibliography}
\end{document}